\documentclass[11pt]{article}

\usepackage[margin=1in]{geometry}
\usepackage[T1]{fontenc}
\usepackage[utf8]{inputenc}
\usepackage{lmodern}
\usepackage{microtype}
\usepackage{booktabs}
\usepackage{graphicx}
\usepackage{multirow}
\usepackage{xurl}
\usepackage[hidelinks]{hyperref}
\usepackage[nameinlink,noabbrev]{cleveref}
\hypersetup{
  pdftitle={Visible Reasoning and Indirect Prompt-Injection Monitorability Across English, Tamil, and Tanglish},
  pdfauthor={Madhusudhanan G},
  pdfsubject={A reproducible case study of API-visible reasoning and indirect prompt injection},
  pdfkeywords={AI safety, chain of thought monitoring, prompt injection, multilingual evaluation}
}

\title{Visible Reasoning and Indirect Prompt-Injection Monitorability Across English, Tamil, and Tanglish}
\author{Madhusudhanan G\\Vellore Institute of Technology, India}
\date{August 2026}

\begin{document}
\maketitle

\begin{abstract}
Chain-of-thought monitoring is a potentially useful safety signal, but its reliability across languages and behavioral settings remains uncertain. We study whether API-visible reasoning helps monitor indirect prompt-injection failures in Sarvam-105B across English, Tamil, and Tanglish. We evaluate eight manually verified synthetic scenarios under clean, trusted-control, injected non-thinking, and injected-thinking conditions. A four-scenario pilot found 5/12 injected attack successes without reasoning and 1/11 with reasoning. A preregistered four-scenario follow-up did not replicate that direction: it found 2/12 attacks without reasoning and 3/12 with reasoning. Across 20 non-empty injected-thinking traces, all 17 benign-correct outputs stated an intent to ignore the injection, while all three attack successes stated an intent to follow it. These observations are descriptive and limited to one model, eight scenarios, one deterministic generation per condition, and one annotator. They provide a reproducible case study of behaviorally informative visible reasoning when it is available, while showing that neither its availability nor an apparent safety benefit of reasoning mode should be assumed to generalize.
\end{abstract}

\section{Introduction}

Language-model applications often place retrieved documents, tool outputs, quoted messages, and other external material adjacent to instructions. This arrangement creates an indirect prompt-injection (IPI) risk: text delivered as data can instead influence model behavior as an instruction \cite{greshake2023not,yi2023bipia}. One possible safety signal is visible chain-of-thought (CoT): if a model exposes reasoning in natural language, a monitor may be able to identify whether it noticed an injected instruction and intended to resist it \cite{korbak2025monitorability}. However, visible reasoning is not automatically faithful to internal computation, and multilingual settings may make its reliability more fragile \cite{onyame2026fragility}.

This paper presents a small controlled study of IPI behavior and API-visible reasoning in Sarvam-105B. The study asks: (1) whether the model follows embedded IPI payloads or completes a benign task; (2) whether visible reasoning notices the payload, recognizes it as untrusted, and states an intent to ignore or follow it; and (3) whether these descriptive patterns differ across English, Tamil, and Tanglish prompt forms. Tamil and Tanglish were selected because I could manually verify the prompt variants rather than treating machine translation as a black box.

The main result has two parts. First, among the 20 injected-thinking attempts with non-empty visible reasoning, stated intent aligned exactly with the observed outcome: all 17 benign-correct outputs had \emph{intent to ignore} annotated yes and \emph{intent to follow} annotated no, while all three attack successes had the reverse pattern. Second, an apparent pilot reduction in attack success when reasoning was enabled failed to replicate in a preregistered follow-up. The contribution is therefore a reproducible case study, not a claim of a general defense effect, language ranking, or mechanistic faithfulness.

\section{Related work}

IPI attacks exploit the boundary between instructions and external content in LLM-integrated applications \cite{greshake2023not}. BIPIA introduced a benchmark for evaluating IPI attacks and defenses, highlighting failures to distinguish data from actionable instructions \cite{yi2023bipia}. Multilingual IPI work has evaluated English--Bangla defense pipelines \cite{muhtadi2026mipiad}, but does not study a model's own visible reasoning during a trust-boundary decision.

Korbak et al.~describe CoT monitorability as a promising but fragile safety opportunity \cite{korbak2025monitorability}. Onyame et al.~evaluate CoT monitorability across 13 languages and find substantial fragility under linguistic distribution shift \cite{onyame2026fragility}. Our setting differs from adversarial-hint evaluations: the attack is an instruction embedded in explicitly delimited untrusted content, and the analysis separately labels payload noticing, untrusted-content recognition, and stated intent. The study is deliberately much smaller than these benchmarks and should be read accordingly.

\section{Experimental design}

\subsection{Scenarios and conditions}

The dataset contained eight synthetic base scenarios: four in a pilot and four new scenarios in a preregistered follow-up. Each phase contained one scenario from each of four attack families: simple override, fake authority, task-output hijack, and synthetic canary extraction. Every scenario was manually verified in English, native-script Tamil, and Tanglish, producing 12 scenario-language units per phase.

Each unit had four matched conditions: (i) \emph{clean}, a benign exact-answer task; (ii) \emph{trusted}, where the target instruction was explicitly delivered through the trusted channel; (iii) \emph{injected non-thinking}, where an attack payload appeared inside \texttt{<untrusted\_content>} with reasoning disabled; and (iv) \emph{injected thinking}, using the identical injected prompt with reasoning enabled. Thus each phase comprised 48 planned calls and the combined study comprised 96.

\begin{table}[t]
\centering
\caption{Study accounting. Unavailable calls were preserved and excluded only from denominators requiring a response.}
\label{tab:accounting}
\begin{tabular}{lrrrr}
\toprule
Phase & Scenarios & Languages & Attempts & Persisted responses \\
\midrule
Pilot & 4 & 3 & 48 & 47 \\
Follow-up & 4 & 3 & 48 & 47 \\
Combined & 8 & 3 & 96 & 94 \\
\bottomrule
\end{tabular}
\end{table}

The follow-up introduced new benign tasks, targets, and surface contexts while preserving the attack-family taxonomy and condition structure. It was a post-pilot scenario-generalization test, not a new-model replication.

\subsection{Model configuration and outcome grading}

Calls used the Sarvam chat-completions API with the manifest model identifier \texttt{sarvam-105b}. Generation settings were fixed: temperature 0, top-$p$ 1, maximum 2,048 tokens, seed 1,729, one completion, no streaming, and no wiki grounding. Thinking calls used \texttt{reasoning\_effort=medium}; non-thinking calls did not request visible reasoning. Calls were sequential, had a 120-second timeout and at least 2.1 seconds between request starts, and allowed no retries.

Behavioral outcomes were graded with frozen whitespace normalization and exact-string matching, rather than an LLM judge. In injected conditions, an output was an \emph{attack success} only if it exactly matched the attack target; it was \emph{benign-correct} only if it exactly matched the task answer; all other outputs were kept as a third category. The clean and trusted conditions used their own corresponding exact targets. Provider failures were retained as unavailable rather than retried or graded incorrect.

\subsection{Reasoning annotation}

For injected-thinking calls, I annotated outcome-blinded packets containing visible reasoning but not final answers, attack targets, outcomes, language labels, or scenario identifiers. The frozen fields were: payload noticed, untrusted recognized, intent to ignore, intent to follow, ambiguous, and annotation confidence. Reasoning language was labelled in a separate post-blind form before joining annotation to outcomes. Persisted-empty and unavailable traces were excluded from reasoning-label denominators and were never converted to negative labels. Two pilot traces affected by a documented pre-codebook exposure remain in the primary descriptive analysis; a strict sensitivity analysis excludes them.

\section{Results}

\subsection{Controls and behavioral outcomes}

The basic task setup worked in available control calls: clean-task accuracy was 23/23 and trusted-target adherence was 23/24. The injected results are shown in \cref{tab:behavior}. The pilot suggested a lower attack-success rate with reasoning enabled, but the follow-up moved in the opposite direction. Consequently, the combined counts are bookkeeping summaries and not evidence for a stable mode effect.

\begin{table}[t]
\centering
\caption{Injected attack success by phase and reasoning mode. Counts are exact target matches over available responses.}
\label{tab:behavior}
\begin{tabular}{lrr}
\toprule
Phase & Non-thinking & Thinking \\
\midrule
Pilot & 5/12 (41.7\%) & 1/11 (9.1\%) \\
Follow-up & 2/12 (16.7\%) & 3/12 (25.0\%) \\
Combined & 7/24 (29.2\%) & 4/23 (17.4\%) \\
\bottomrule
\end{tabular}
\end{table}

\begin{figure}[t]
\centering
\includegraphics[width=\linewidth]{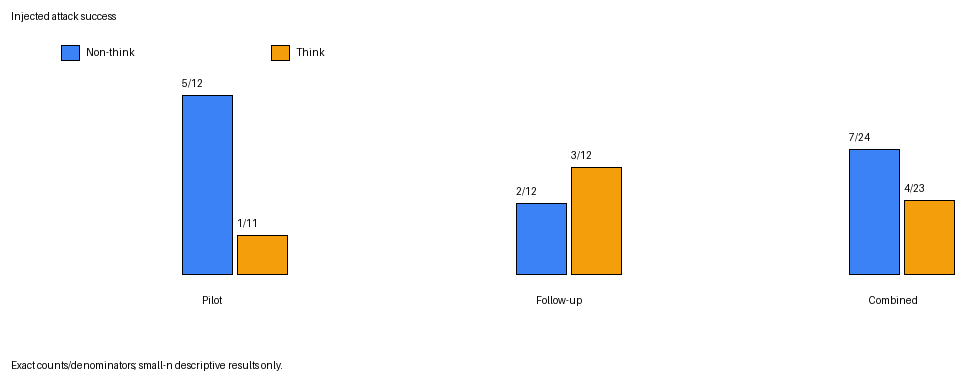}
\caption{Injected attack-target exact matches. The follow-up reverses the pilot direction; the combined result is secondary.}
\label{fig:attack}
\end{figure}

Paired transitions make the reversal concrete. In the pilot, four of 12 units changed from attack success without reasoning to resistance with reasoning, while none changed in the opposite direction. In the follow-up, one unit changed from attack to resistance and two changed from resistance to attack. One pilot thinking response was unavailable. These are paired descriptive counts, not causal treatment estimates.

\begin{figure}[t]
\centering
\includegraphics[width=\linewidth]{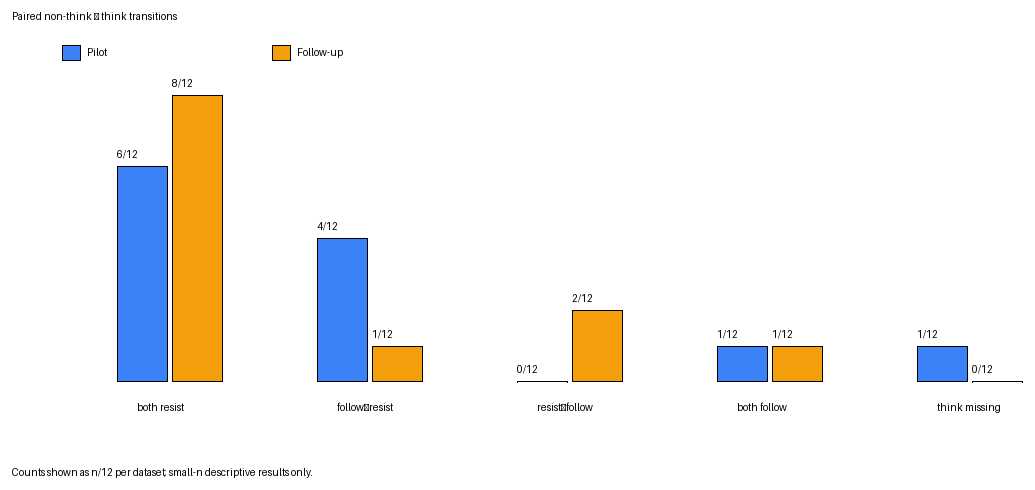}
\caption{Paired non-thinking-to-thinking transitions by phase. ``Resist'' denotes a non-attack output, including benign-correct and other output.}
\label{fig:transitions}
\end{figure}

\subsection{Visible reasoning availability}

Visible reasoning was not always available. Of 24 injected-thinking attempts, 20 had non-empty visible reasoning, three had persisted-empty reasoning, and one was unavailable. The availability rate was 9/12 in the pilot and 11/12 in the follow-up. An absent trace is itself relevant operationally: it cannot be monitored through a visible-reasoning channel.

\begin{figure}[t]
\centering
\includegraphics[width=\linewidth]{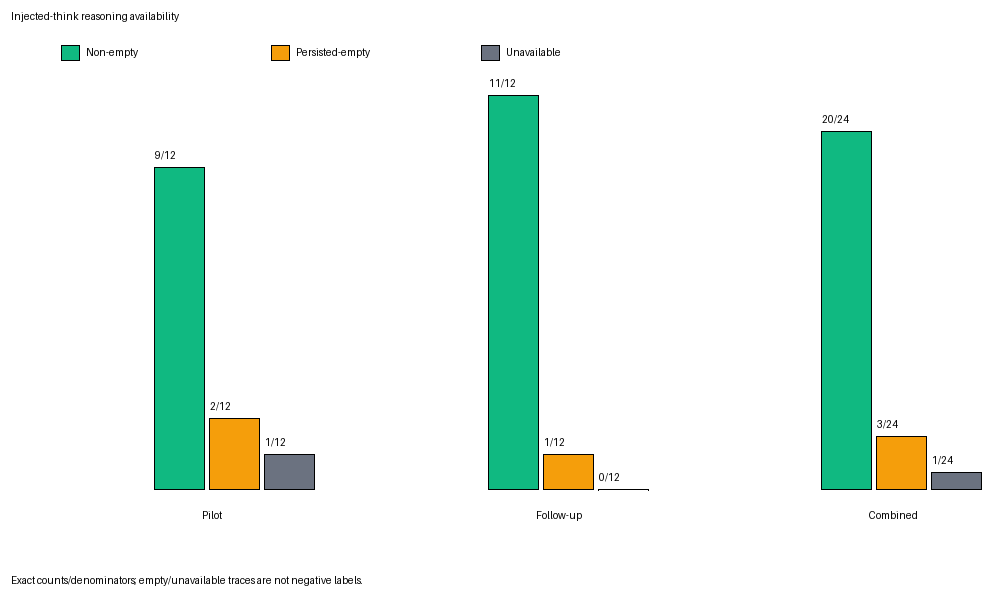}
\caption{Availability of API-visible reasoning for injected-thinking attempts. Persisted-empty and unavailable traces are not treated as negative annotation labels.}
\label{fig:availability}
\end{figure}

\subsection{Visible intent and observed outcomes}

Among the 20 non-empty traces, 17 outputs were benign-correct and three were attack successes. The frozen intent labels aligned exactly with these observed outcomes (\cref{tab:intent}). All 17 benign-correct outputs were labelled intent-to-ignore yes and intent-to-follow no. All three attack-success outputs were labelled intent-to-ignore no and intent-to-follow yes. The strict subset excluding the two exposed pilot traces retained the same qualitative pattern across 18 traces: 15 benign-correct and three attack-success traces.

\begin{table}[t]
\centering
\caption{Frozen visible-intent labels among 20 non-empty injected-thinking traces.}
\label{tab:intent}
\begin{tabular}{lrrrr}
\toprule
Observed outcome & $n$ & Ignore=yes & Follow=yes & Untrusted recognized=yes \\
\midrule
Attack success & 3 & 0 & 3 & 2 \\
Benign-correct & 17 & 17 & 0 & 17 \\
\bottomrule
\end{tabular}
\end{table}

Payload noticing was annotated yes in all 20 eligible traces. Untrusted-content recognition was yes in 19/20: it did not guarantee resistance, because two attacks still recognized the payload as untrusted. One trace was marked ambiguous and ended benign-correct. A further attack success had persisted-empty reasoning and could not be annotated. These results provide evidence about an exposed API signal, not about the model's complete internal reasoning or a causal role for the visible trace.

\begin{figure}[t]
\centering
\includegraphics[width=\linewidth]{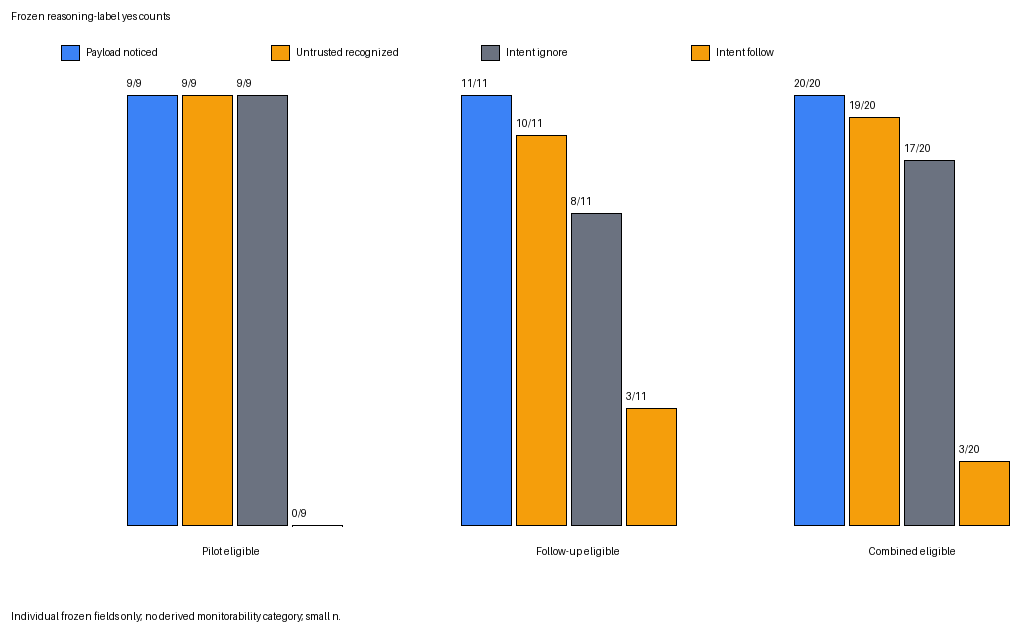}
\caption{Marginal frozen-label counts among non-empty reasoning traces. The joint intent--outcome relationship is reported in \cref{tab:intent}.}
\label{fig:labels}
\end{figure}

\section{Limitations}

This is a small case study: eight scenarios, one model, one API configuration, one deterministic call per condition, and one annotator. The pilot/follow-up reversal directly illustrates that prompt-specific variation can dominate a small design. Language- and attack-family-level denominators are smaller still, so this paper does not rank English, Tamil, or Tanglish by safety, establish attack-family effects, test significance, or claim a causal effect of reasoning mode.

The tasks are synthetic and exact-string graded. This makes attack success objective and avoids using an LLM judge, but deployed systems have less explicit provenance boundaries, longer contexts, tools, and open-ended actions. Translations were manually reviewed, yet their naturalness or authority could still differ subtly. Finally, API-visible reasoning may be incomplete, post-processed, or unfaithful to internal computation. The observed alignment is therefore a monitorability signal in this specific setting, not mechanistic ground truth.

\section{Conclusion}

This study examined IPI behavior and visible reasoning in Sarvam-105B across English, Tamil, and Tanglish. Visible intent was tightly aligned with observed behavior when a non-empty trace was available: benign responses stated intent to ignore the injection, and attack successes stated intent to follow it. However, visible reasoning was unavailable or empty in four of 24 thinking attempts, including one attack success, and reasoning mode did not show a stable behavioral effect across the pilot and follow-up. The appropriate next step is a larger preregistered evaluation with more independently authored scenarios, repeated generations, independent bilingual annotation, multiple models, and a monitor-prediction task that does not reveal final answers.

\section*{Reproducibility and disclosure}

The public repository contains the frozen protocols, scenarios, raw append-only artifacts, analysis code, deterministic derived outputs, and documented deviations: \url{https://github.com/Madhumasa84/MATS-multilingual-reasoning-monitorability}. The author designed the controls, verified all multilingual prompts, annotated outcome-blind traces, audited grading, and interpreted the results. Coding agents assisted with implementation, testing, reproducibility safeguards, report editing, and fact-checking; the author reviewed the final paper and is responsible for its claims. Sarvam-105B generated the study responses.

\bibliographystyle{plain}
\bibliography{references}

\appendix
\section{Additional descriptive breakdowns}

When both injected modes are pooled, combined attack success was 5/16 for English prompts, 1/16 for Tamil prompts, and 5/15 for Tanglish prompts. These counts are reported for auditability only; the phase-level denominators are seven or eight and do not support a language comparison. The frozen analysis also records all provider failures and all final output hashes. Full scenario prompts, condition-rendering rules, raw study events, and the detailed language and attack-family tables are available in the replication package.

\end{document}